\documentclass[letterpaper]{article} 
\pdfoutput=1 
\usepackage[preprint]{aaai2027}  
\usepackage[hyphens]{url}  
\usepackage{graphicx} 
\usepackage{natbib}  
\usepackage{caption} 
\usepackage{algorithm}
\usepackage{algorithmic}
\usepackage{newfloat}
\usepackage{listings}
\DeclareCaptionStyle{ruled}{labelfont=normalfont,labelsep=colon,strut=off} 
\floatstyle{ruled}
\newfloat{listing}{tb}{lst}{}
\floatname{listing}{Listing}
\usepackage{booktabs}
\usepackage{xcolor}
\usepackage{multirow}
\usepackage{amsmath}
\usepackage{amssymb}
\usepackage{colortbl}
\definecolor{bandcol}{gray}{0.94}
\definecolor{deltacol}{rgb}{0.17,0.32,0.75}
\definecolor{citecol}{rgb}{0.11,0.33,0.60}
\definecolor{refcol}{rgb}{0.58,0.14,0.20}
\makeatletter
\def\hyper@natlinkstart#1{\begingroup\color{citecol}}
\def\hyper@natlinkend{\endgroup}
\makeatother
\newcommand{\xref}[1]{\textcolor{refcol}{#1}}
\newcommand{\figref}[1]{\xref{Figure~\ref{#1}}}

\newcommand{\tabref}[1]{\xref{Table~\ref{#1}}}
\newcommand{\subtabref}[2]{\xref{Table~\ref{#1}(#2)}}
\newcommand{\secref}[1]{\xref{\S\ref{#1}}}
\newcommand{\equref}[1]{\xref{Eq.~\eqref{#1}}}
\newcommand{\dgain}[1]{\textcolor{deltacol}{$\uparrow$#1}}
\newcommand{\dloss}[1]{\textcolor{deltacol}{$\downarrow$#1}}
\newcommand{\venue}[1]{$_{\text{\tiny #1}}$}
\newcommand{\cmfcell}[2]{%
  \shortstack[r]{%
    \textbf{#1}\\[-0.65ex]
    {\scriptsize\dgain{#2}}%
  }%
}

\title{Contrastive Mask Fidelity: Reference-Free Auditing of Ground-Truth Masks in Remote Sensing Semantic Segmentation}

\author{
    Shuaishuai Cao\textsuperscript{\rm 1},
    Shuwei Peng\textsuperscript{\rm 2},
    Meng Tang\textsuperscript{\rm 3},
    Min Huang\textsuperscript{\rm 4,\dag},\\
    Youjin Wang\textsuperscript{\rm 5},
    Jie Chen\textsuperscript{\rm 1},
    Jing Ouyang\textsuperscript{\rm 4},
    Zhiwei Zhai\textsuperscript{\rm 6}
}
\affiliations{
    \textsuperscript{\rm 1}Central South University \quad
    \textsuperscript{\rm 2}University of Chinese Academy of Sciences \quad
    \textsuperscript{\rm 3}Aberystwyth University\\
    \textsuperscript{\rm 4}Jiangxi Normal University \quad
    \textsuperscript{\rm 5}Renmin University of China \quad
    \textsuperscript{\rm 6}BGI Research\\[2pt]
    265011218@csu.edu.cn \quad huangm@jxnu.edu.cn\\[2pt]
    \textsuperscript{\rm \dag}Corresponding author
}

\begin{document}

\maketitle

\begin{abstract}
Semantic segmentation models are trained and evaluated against human-drawn
masks, yet remote-sensing annotations are often coarse, incomplete, or
misaligned; high overlap scores may then reflect agreement with imperfect
labels rather than faithfulness to the image---an evaluation paradox.
We introduce \emph{Contrastive Mask Fidelity} (CMF), a training-free,
reference-free metric that scores competing class masks directly against
image evidence: CMF composites keep and erase counterfactual views of each
mask and asks a frozen vision-language judge whether class evidence is
concentrated inside the mask and absent outside.
We validate CMF on controlled mask corruptions, then audit $10{,}731$
image-class pairs across ten remote-sensing benchmarks using candidate
masks from \emph{Seg-Probe}, a training-free open-vocabulary probe built
on SegEarth-OV3 that outperforms prior baselines on nine of ten datasets.
The audit reveals systematic, class-dependent annotation distortion:
man-made classes such as buildings, roads, and cars favor the candidate
mask on $62$--$85\%$ of pairs, whereas ambiguous land cover more often
favors human annotations.
On a blinded three-annotator consensus, CMF matches expert judgment on
$81\%$ of pairs, exceeding keep-only scoring, model confidence, and a
trained label-quality baseline.
Finally, conservative class-wise arbitration yields supervision that
improves cross-domain transfer over raw annotations and matched
replacement controls, positioning CMF as a scalable tool for auditing
ground truth rather than presuming it infallible.
\end{abstract}

\section{Introduction}

\begin{figure}[t]
\centering
\includegraphics[width=1.0\linewidth]{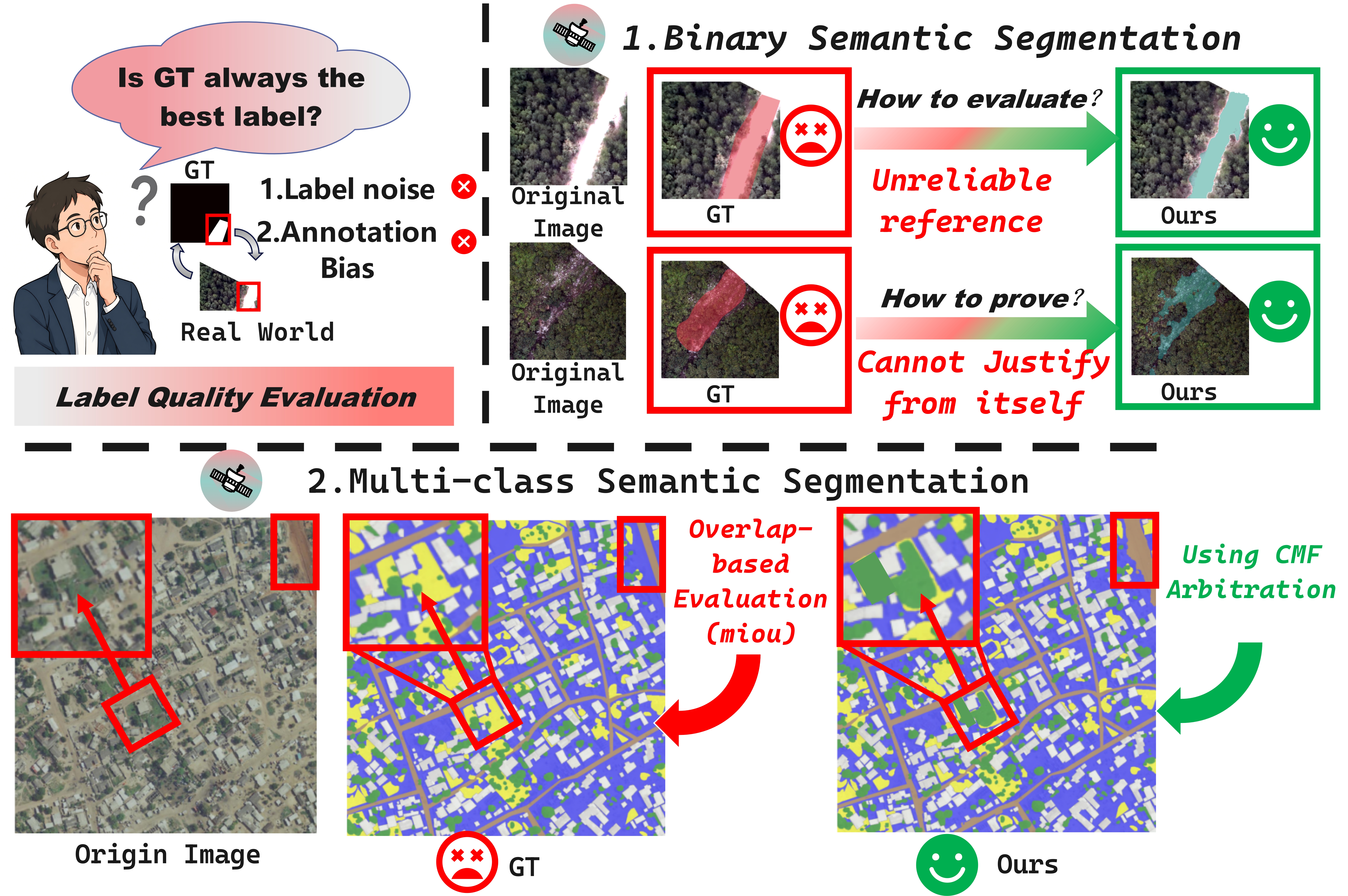}
\caption{\textbf{Is the ground truth always the best label?}
On binary (\emph{top}) and multi-class (\emph{bottom}) benchmarks, the human
ground truth misses or distorts the true extent while a training-free
prediction follows the image. How do we \emph{evaluate}---and
\emph{exploit}---mask fidelity without trusting the annotation?}
\label{fig:teaser}
\end{figure}

Semantic segmentation is evaluated as if the reference mask were exact.
Overlap metrics such as mIoU therefore quantify agreement with an annotation,
not necessarily faithfulness to the image: when the annotation is incomplete,
misregistered, or over-smoothed, a visually better mask can receive a worse
score.
We call this the \emph{evaluation paradox}, and it is rarely questioned because
the reference is, by convention, called ground truth.

Remote sensing amplifies the paradox.
Aerial and satellite scenes are densely packed with small, ambiguous, and
heterogeneous objects; polygonized labels smooth boundaries, imagery and label
sources can be misaligned, and thin structures are easily omitted
(\figref{fig:teaser}).
Prior work has shown that even flagship classification benchmarks carry enough
label errors to reorder model rankings~\citep{northcutt2021labelerrors}, and
label-quality scores for segmentation confirm that pixel-accurate annotation
is error-prone at scale~\citep{lad2023segquality}.

Existing tools do not resolve the paradox.
Overlap metrics presuppose a trusted reference.
Label-noise and label-quality methods score suspicious labels with a model
trained on the same label distribution, so they can inherit the very biases
they are meant to detect~\citep{lad2023segquality,li2023noisyseg}.
Strong promptable foundation models such as SAM\,3~\citep{sam3} and
training-free open-vocabulary segmenters for remote
sensing~\citep{segearthov,segearthov3} now produce masks that visibly disagree
with human labels, but a high-quality prediction is a \emph{candidate}, not a
verdict.
What is missing is an adjudicator that compares two competing masks against
the image itself, without designating either as the reference.

We argue that such an adjudicator must satisfy four requirements:
it should treat the two competing masks \emph{symmetrically}; it should be
\emph{image-conditioned} rather than reference-conditioned; it should test both
\emph{purity} (the claimed region depicts the class) and \emph{completeness}
(no class evidence remains outside); and it should be \emph{external and
frozen}, with parameters independent of the segmenter under audit.
\emph{Contrastive Mask Fidelity} (CMF) meets these requirements by contrasting
two counterfactual views: a \emph{keep} view that reveals only the masked
region, and an \emph{erase} view that hides it, both scored by a frozen
remote-sensing vision-language judge (RemoteCLIP) through a softmax over the
dataset's class texts.
The candidate masks are supplied by \emph{Seg-Probe}, a strong training-free
candidate generator built on SegEarth-OV3~\citep{segearthov3} with synonym
evidence pooling, multi-scale evidence, and test-time augmentation
(\tabref{tab:model}).
Seg-Probe is deliberately \emph{not} a contribution we build the paper on:
its role is to expose mask disagreements strong enough to warrant
adjudication; we summarize its design in \secref{sec:segprobe} and defer
derivations to the supplementary material.
We call the complete three-stage auditing system \emph{GeoVeritas}
(\figref{fig:pipeline}).

Our evidence chain has four links.
First, we \emph{calibrate} CMF on controlled mask corruptions and analyze its
residual sensitivity to mask area, prompts, and judge choice
(\secref{sec:validity}, \secref{sec:ablation}).
Second, we \emph{audit} $10{,}731$ image-class pairs across ten benchmarks,
revealing class-structured disagreement: well-delineated man-made classes
often favor the model mask, amorphous land cover favors the human annotation
(\secref{sec:audit}).
Third, we \emph{ground} CMF in expert judgment: on a stratified disagreement
set with blinded three-annotator majority votes, CMF matches the expert
majority more often than keep-only VLM scoring, model confidence, and a
trained label-quality baseline (\secref{sec:human}).
Fourth, we test whether verdicts are \emph{actionable and safe}: a
lightweight segmenter trained on labels arbitrated by CMF verdicts
transfers better across domains than models trained on the raw
annotations, on pure predictions, or on matched random- and
confidence-replacement controls, isolating the arbitration effect from
pseudo-label distillation
(\secref{sec:crossdomain}, \figref{fig:crossviz}).

\noindent\textbf{Contributions.}
\begin{itemize}
\item We formulate \textbf{CMF}, a reference-free mask-fidelity metric that
turns mask disagreement into an image-conditioned keep--erase contrast,
designed to reduce area sensitivity, with a frozen external vision-language
judge.
\item We conduct an \textbf{expert-grounded audit} of $10{,}731$ image-class
pairs across ten remote-sensing datasets: CMF is calibrated on corruption
controls, tracks a blinded expert consensus more closely than three simpler
adjudication rules, and reveals class-structured annotation
distortion.
\item We show the audit supports \textbf{trustworthy correction}:
conservative class-wise arbitration of the verdicts improves cross-domain
transfer, and matched distillation controls show the gain comes from CMF's
selection rather than from strong pseudo-labels alone.
\end{itemize}

\section{Related Work}
\label{sec:related}

\paragraph{Training-free open-vocabulary segmentation.}
SAM~\citep{sam} established promptable class-agnostic segmentation, extended to
video by SAM\,2~\citep{sam2} and to exhaustive open-vocabulary concept
segmentation by SAM\,3~\citep{sam3}.
Training-free CLIP adaptations such as MaskCLIP~\citep{maskclip},
SCLIP~\citep{sclip}, GEM~\citep{gem}, ClearCLIP~\citep{clearclip},
ProxyCLIP~\citep{proxyclip}, and CorrCLIP~\citep{corrclip} improve dense
localization from pretrained vision-language models.
In remote sensing, SegEarth-OV~\citep{segearthov} adds feature upsampling and
global-bias debiasing; \citet{ovrs} and GSNet~\citep{gsnet} tailor
open-vocabulary segmentation to aerial orientation and scale variation;
ReAttnCLIP~\citep{reattnclip} re-defines CLIP attention
for aerial scenes; ConInfer~\citep{coninfer} adds context-aware joint
inference; and SegEarth-OV3~\citep{segearthov3} explores SAM\,3's dual heads
and presence filtering.
All of these works optimize \emph{candidate mask generation}; none asks how to
adjudicate a disagreement between a candidate and the human annotation without
trusting the latter.
Seg-Probe belongs to this family and is compared against it in
\tabref{tab:model}, but in this paper it serves as a candidate source for
the audit.

\paragraph{Segmentation label quality and label noise.}
\citet{northcutt2021labelerrors} showed pervasive label errors in
classification benchmarks.
For segmentation, \citet{lad2023segquality} score per-image label quality from
the predictions of a model trained on the audited label distribution;
SegAssess~\citep{segassess} trains a SAM-based network to predict pixel-wise
quality maps for a given mask; AIO2~\citep{aio2} corrects incomplete object
labels online during training; and \citet{li2023noisyseg} learn robustly
\emph{despite} noisy masks.
All of these rely on segmentation models trained on (or corrected against)
the same label distribution, so they can inherit annotation bias.
CMF instead compares the two competing masks through an external
image--text model trained for alignment rather than segmentation, requires
no training, and does not designate either mask as the reference.

\paragraph{Vision-language models as evaluators.}
CLIP~\citep{clip} enables reference-free image--text scoring, popularized by
CLIPScore~\citep{clipscore}; RemoteCLIP~\citep{remoteclip} specializes the
capability for remote sensing.
Recent work also trains generative VLM judges with score
rubrics~\citep{prometheusvision}, whose reliability is itself an active
research question.
Unlike rubric-based holistic ratings, CMF uses a fixed, interpretable
keep--erase contrast; we treat judge reliability as an empirical question and
verify that audit conclusions are stable across three alternative judges
(\secref{sec:ablation}).

\section{Method}
\label{sec:method}

\begin{figure*}[t]
\centering
\includegraphics[width=0.97\textwidth]{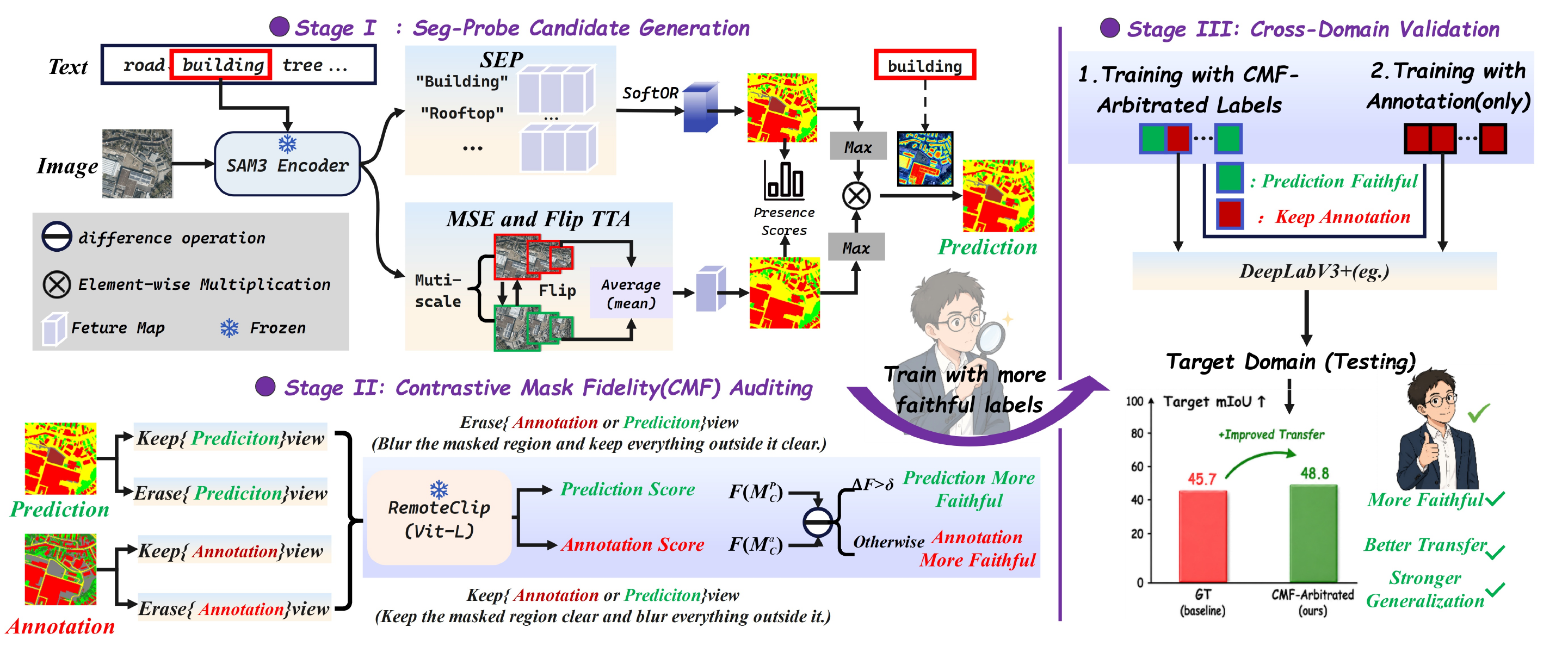}
\caption{\textbf{The GeoVeritas pipeline.}
\emph{Stage I}: Seg-Probe generates candidate masks with a frozen SAM\,3
encoder, pooling synonym prompts by a soft-OR (SEP) and fusing multi-scale
and flipped views (MSE, TTA).
\emph{Stage II}: for each class, CMF builds keep/erase views of both the
prediction and the ground truth and lets a frozen RemoteCLIP judge decide
which mask is more faithful ($\Delta F$).
\emph{Stage III}: labels arbitrated by the verdicts train a segmenter that is
tested on unseen target domains.}
\label{fig:pipeline}
\end{figure*}

GeoVeritas is designed to assess whether the human annotation or the model
prediction more faithfully reflects the real-world scene depicted in the
image, without treating either as an unquestioned reference. It audits an
annotated dataset in three stages
(\figref{fig:pipeline}):
\emph{Stage I}---\emph{Seg-Probe} produces a candidate mask for each image
(\secref{sec:segprobe});
\emph{Stage II}---\emph{CMF} scores the candidate and the human annotation
with the same frozen judge (\secref{sec:cmf}) and arbitrates per class
(\secref{sec:arbitration});
\emph{Stage III}---the arbitrated labels are validated by cross-domain
training (\secref{sec:crossdomain}).

\subsection{Problem Formulation}
\label{sec:prelim}

Let $x\in\mathbb{R}^{H\times W\times 3}$ denote an input image, and let
$\mathcal{C}=\{c_1,\dots,c_K\}$ denote the set of $K$ semantic classes in the
dataset. A semantic segmentation map is represented as
$Y\in\{1,\dots,K\}^{H\times W}$, where each pixel is assigned to one class.
When a dataset defines background as a semantic category, background is
included in $\mathcal{C}$.
Let $Y^{a}$ denote the human annotation and let $Y^{p}$ denote the
segmentation predicted by Seg-Probe.
For a segmentation source, the binary mask of class $c$ is defined as
\begin{equation}
M_c = \mathbf{1}\left[Y=c\right],
\label{eq:class_mask}
\end{equation}
where $\mathbf{1}[\cdot]$ denotes the indicator function.
GeoVeritas audits each semantic class in an image separately. Specifically,
for each image--class pair $(x,c)$, it compares the human-annotation mask
$M_c^{a}$ with the Seg-Probe mask $M_c^{p}$.

The human annotation and the Seg-Probe prediction therefore
induce two sets of class masks,
$\{M_c^{a}\}_{c\in\mathcal{C}}$ and
$\{M_c^{p}\}_{c\in\mathcal{C}}$.
Within each segmentation map, the masks are mutually
exclusive and jointly describe the valid image region.
For each image, we audit only the classes present in its human annotation,
those satisfying
\[
\mathcal{A}(x)
= \left\{ c\in\mathcal{C}:
\left\lVert M_c^{a}\right\rVert_1>0 \right\}.
\]
For each $c\in\mathcal{A}(x)$, GeoVeritas compares $M_c^{a}$ and $M_c^{p}$
using the same input image, class vocabulary, text prompts, masking procedure,
and evaluator. An empty Seg-Probe mask remains an audit input because it
represents a complete miss of an annotated class. Void pixels are excluded
from both mask evaluations. Classes appearing only in the Seg-Probe prediction
are outside the audit scope and are instead accounted for by the standard mIoU
evaluation reported in \tabref{tab:model}.

\subsection{Seg-Probe: Training-Free Candidate Generation}
\label{sec:segprobe}

Seg-Probe generates a candidate segmentation $Y^{p}$ for subsequent
comparison with the human annotation
(\figref{fig:pipeline}, Stage~I).
It builds on SegEarth-OV3~\citep{segearthov3}, which uses a frozen
SAM\,3 encoder~\citep{sam3}, and aggregates class evidence through
three training-free components.

First, synonym evidence pooling (SEP) queries multiple synonymous
prompts for each class and combines their evidence maps using a
log-sum-exp soft-OR, reducing reliance on any single class name.
Second, multi-scale evidence (MSE) extracts complementary evidence
from two image scales and aligns the resulting maps to the original
resolution. Third, horizontal-flip test-time augmentation averages
the evidence from the original and flipped views after spatial
realignment. The SEP and MSE/TTA evidence streams are fused by an
element-wise maximum and subsequently modulated by the corresponding
class-presence scores. Per-pixel class assignment then produces
$Y^{p}$.

All model parameters remain frozen, and the module configuration is
fixed before CMF auditing. Exact prompts, scales, pooling details,
and dataset-level activation settings are provided in the
supplementary material. Seg-Probe serves only as a strong candidate
generator; CMF is agnostic to the particular segmentation backbone.

\subsection{Contrastive Mask Fidelity}
\label{sec:cmf}
Given an image $x$ and a binary class mask $M_c$, CMF asks whether
$M_c$ faithfully delineates where class $c$ appears in $x$---a property we
term \emph{semantic mask fidelity}. Answering this question without trusting
any annotation requires an external judge that measures the agreement between
an image region and a class name, together with a mechanism for interrogating
that judge about one specific mask. We introduce both below.

\paragraph{A frozen external judge.}
We require a function that, given an image region and a class name, returns
how strongly the region supports that class relative to the other classes, and
whose parameters are independent of the segmenter under audit.
Remote-sensing vision-language models provide a practical source of such
class-relative evidence: they align aerial imagery with text in a shared
embedding space and support zero-shot classification from natural language.
Following the reference-free evaluation tradition of
CLIPScore~\citep{clipscore}, we adopt RemoteCLIP~\citep{remoteclip} as the
judge, with image encoder $\phi(\cdot)$ and text encoder $\psi(\cdot)$.
The judge is frozen throughout, is optimized for image--text alignment rather
than mask overlap, and shares no parameters with SAM\,3, so both candidates
are scored on equal footing; \secref{sec:ablation} further verifies that the
audit conclusions are stable under three alternative judges.
For any view $r$ of $x$, the class-conditional belief
of the judge is a temperature-scaled softmax over the class texts of the dataset,
\begin{equation}
  P(c \mid r)
  = \frac{\exp\!\big(\tau \,\langle \phi(r), \bar{\psi}(c) \rangle\big)}
         {\sum_{k=1}^{K} \exp\!\big(\tau \,\langle \phi(r), \bar{\psi}(c_k) \rangle\big)},
  \label{eq:belief}
\end{equation}
where $\langle \cdot, \cdot \rangle$ denotes cosine similarity, $\tau$
is a fixed logit scale, and $\bar{\psi}(c)$ is a prompt-ensembled text
embedding, that is, the renormalized average of $\psi$ over synonym phrases of class $c$. Because the softmax measures relative class
evidence, it is far less sensitive than raw similarity to absolute embedding magnitude, which grows with region area. We exploit this
property below and quantify the residual sensitivity in \secref{sec:validity}.

\paragraph{Counterfactual views.}
To interrogate the judge about $M_c$, we composite two counterfactual
views of $x$, namely the keep view $x^{k}$ and the erase view
$x^{e}$, with a Gaussian blur $B(\cdot)$, which suppresses content
while preserving local color statistics and avoiding
out-of-distribution black fills:
\begin{align}
  x^{k}_c  &= x \odot M_c + B(x) \odot (1 - M_c),
  \label{eq:keep}\\
  x^{e}_c &= x \odot (1 - M_c) + B(x) \odot M_c,
  \label{eq:erase}
\end{align}
where $\odot$ denotes pixel-wise multiplication. The keep view
reveals only what $M_c$ claims to be class $c$, whereas the erase
view reveals everything else.
 
\paragraph{Fidelity score.}
Contrastive Mask Fidelity measures how much more the judge believes
in class $c$ when shown the keep view than when shown the erase view:
\begin{equation}
  F(M_c) = P\big(c \mid x^k_c\big)
           - P\big(c \mid x^e_c\big),
  \label{eq:cmf}
\end{equation}
with $F(M_c) \in [-1, 1]$, where higher values indicate a more
faithful mask. A faithful mask is both pure and complete. Purity
means that the kept region depicts class $c$, which drives the first
term up; completeness means that no evidence of $c$ remains outside
$M_c$, which drives the second term down. The contrastive construction is
designed to reduce trivial area inflation:
the softmax in \equref{eq:belief} normalizes away
class-agnostic magnitude growth, and the two terms of
\equref{eq:cmf} respond to mask enlargement in opposite directions, so
enlarging $M_c$ does not automatically inflate $F$.
We measure the residual area sensitivity empirically in
\secref{sec:validity}.
 
\paragraph{Verdict.}
For class $c$ in image $x$, the prediction is judged more faithful
than the human label if and only if
\begin{equation}
  \Delta F = F\big(M^p_c\big)
             - F\big(M^a_c\big) > 0.
  \label{eq:verdict}
\end{equation}
We report the prediction win rate, defined as the fraction of audited
pairs that satisfy \equref{eq:verdict}, at both the class level
and the dataset level.
 
\subsection{From Verdicts to Arbitrated Labels}
\label{sec:arbitration}
 
CMF would be of limited value if it were only diagnostic. If it
correctly identifies distorted annotations, then substituting the preferred
mask should yield more useful supervision---a prediction we test under
cross-domain transfer in \secref{sec:crossdomain}. For each audited pair
$(x, c)$, we arbitrate per class,
\begin{equation}
  \tilde{M}_c =
  \begin{cases}
    M^p_c, & \text{if } \Delta F > \delta,\\[2pt]
    M^a_c,   & \text{otherwise},
  \end{cases}
  \label{eq:arbitration}
\end{equation}
with margin $\delta = 0$ by default, so that ties defer to the human
label. The arbitrated map $\tilde{Y}$ reassembles the family
$\{\tilde{M}_c\}$: a pixel claimed by exactly one arbitrated mask takes that
class, and a pixel claimed by several arbitrated masks, or by none, retains
its human label. All conflicts therefore resolve conservatively in favor of
the ground truth.
The training protocol and the cross-domain evaluation of $\tilde{Y}$ are given
in \secref{sec:crossdomain}.

\section{Experiments}
\label{sec:experiments}

We evaluate GeoVeritas on ten remote-sensing segmentation datasets spanning
$0.05$--$1$\,m GSD and three platforms (satellite, airborne, UAV): a UAV
landslide suite, LoveDA~\citep{loveda}, Potsdam and
Vaihingen~\citep{isprs2d}, OpenEarthMap~\citep{openearthmap},
LandCover.ai~\citep{landcoverai}, DeepGlobe~\citep{deepglobe},
FloodNet~\citep{floodnet}, and Massachusetts Buildings and
Roads~\citep{mnih2013}, yielding $10{,}731$ audited image--class pairs.
Each dataset carries documented annotation limitations---polygonized or
over-smoothed boundaries (LoveDA, Potsdam, DeepGlobe), sparse or incomplete
object tracing (LandCover.ai, Massachusetts Buildings/Roads),
imagery--label misregistration (Massachusetts Buildings), coarse landslide
scars (Landslide), and IRRG false color outside the RGB judge domain
(Vaihingen)---so the audit is exercised on precisely the conditions that make
the evaluation paradox concrete.
Using a frozen RemoteCLIP ViT-L/14 judge and Seg-Probe candidate masks, the
experiments answer four questions in order:
(RQ1) are the candidate masks strong enough that their disagreements with
human labels deserve adjudication (\secref{sec:segresults});
(RQ2) does CMF respond correctly to known corruptions while remaining stable
to mask area (\secref{sec:validity});
(RQ3) what does the audit reveal, and does CMF track independent expert
judgment better than simpler adjudication rules
(\secref{sec:audit}, \secref{sec:human});
(RQ4) are the verdicts actionable as supervision once pseudo-label
distillation is controlled for (\secref{sec:crossdomain})?
Dataset statistics and implementation details are reported in
\tabref{tab:main} and the supplementary material.

\begin{figure*}[t]
\centering
{\scriptsize
\setlength{\tabcolsep}{4.6pt}
\renewcommand{\arraystretch}{1.08}
\begin{tabular}{@{}l cccccccccc c@{}}
\toprule
\textbf{Method} & \textbf{Land.} & \textbf{LoveDA} & \textbf{Pots.} &
\textbf{Vaih.} & \textbf{OEM} & \textbf{LC.ai} & \textbf{DGlobe} &
\textbf{Flood.} & \textbf{M-Bldg} & \textbf{M-Road} & \textbf{Avg.} \\
\midrule
MaskCLIP\venue{ECCV'22}      & 54.2 & 27.8 & 33.9 & 29.9 & 25.1 & 28.4 & 24.6 & 31.8 & 52.3 & 48.6 & 35.7 \\
SCLIP\venue{ECCV'24}         & 57.4 & 30.4 & 39.6 & 35.9 & 29.3 & 31.2 & 27.1 & 34.5 & 55.6 & 51.2 & 39.2 \\
GEM\venue{CVPR'24}           & 59.8 & 31.6 & 39.1 & 36.4 & 33.9 & 32.6 & 28.3 & 36.2 & 57.1 & 52.8 & 40.8 \\
ClearCLIP\venue{ECCV'24}     & 61.1 & 32.4 & 42.0 & 36.2 & 31.0 & 34.1 & 29.8 & 37.4 & 58.9 & 54.3 & 41.7 \\
ProxyCLIP\venue{ECCV'24}     & 62.7 & 34.3 & 49.0 & 47.5 & 38.9 & 35.9 & 31.4 & 39.1 & 60.4 & 56.7 & 45.6 \\
CorrCLIP\venue{ICCV'25}      & 63.5 & 36.9 & 51.9 & 47.0 & 32.9 & 38.0 & 33.0 & 41.2 & 63.1 & 59.0 & 46.7 \\
SegEarth-OV\venue{CVPR'25}   & 64.4 & 36.9 & 48.5 & 40.0 & 40.3 & 37.2 & 32.5 & 40.8 & 62.7 & 58.1 & 46.1 \\
SegEarth-OV3\venue{arXiv'25} & \underline{66.8} & \underline{46.0} & \underline{57.1} & \textbf{62.5} & \underline{44.9} & \underline{61.8} & \underline{36.8} & \underline{41.9} & \underline{68.9} & \underline{66.9} & \underline{55.4} \\
\midrule
(a) Baseline (reproduced)             & 66.8 & 46.0 & 57.1 & 62.5 & 44.9 & 61.8 & 36.8 & 41.9 & 68.9 & 66.9 & 55.4 \\
(b) $+$ Synonym Evidence Pooling      & 66.8 & 46.5 & 57.2 & 62.5 & 46.1 & 62.9 & 37.7 & 42.2 & 69.0 & 67.0 & 55.8 \\
(c) $+$ Multi-Scale Evidence (gated)  & 66.8 & 46.8 & 57.2 & 62.5 & 47.5 & 64.1 & 38.1 & 42.4 & 69.0 & 67.0 & 56.1 \\
\rowcolor{bandcol}
(d) $+$ Flip TTA\enspace(= \textbf{Seg-Probe}) & \textbf{66.9} & \textbf{46.9} & \textbf{57.3} & \underline{62.4} & \textbf{47.9} & \textbf{64.5} & \textbf{38.5} & \textbf{42.5} & \textbf{69.2} & \textbf{67.0} & \textbf{56.3} \\
\rowcolor{bandcol}
\emph{Seg-Probe} ($\Delta$) & \dgain{0.1} & \dgain{0.9} & \dgain{0.2} & \dloss{0.1} & \dgain{3.0} & \dgain{2.7} & \dgain{1.7} & \dgain{0.6} & \dgain{0.3} & \dgain{0.1} & \dgain{0.9} \\
\bottomrule
\end{tabular}
\captionof{table}{\textbf{Candidate quality measured by training-free mIoU
(\%) on ten audited datasets.}
The upper block compares prior methods, and the lower block reports the
cumulative Seg-Probe ablation, with row~(d) denoting the full model.
``Seg-Probe~($\Delta$)'' reports its improvement over SegEarth-OV3.
\textbf{Bold}/\underline{underline} indicate the best/second-best result in
each column.}
\label{tab:model}
}

\vspace{7pt}

\includegraphics[width=0.90\textwidth]{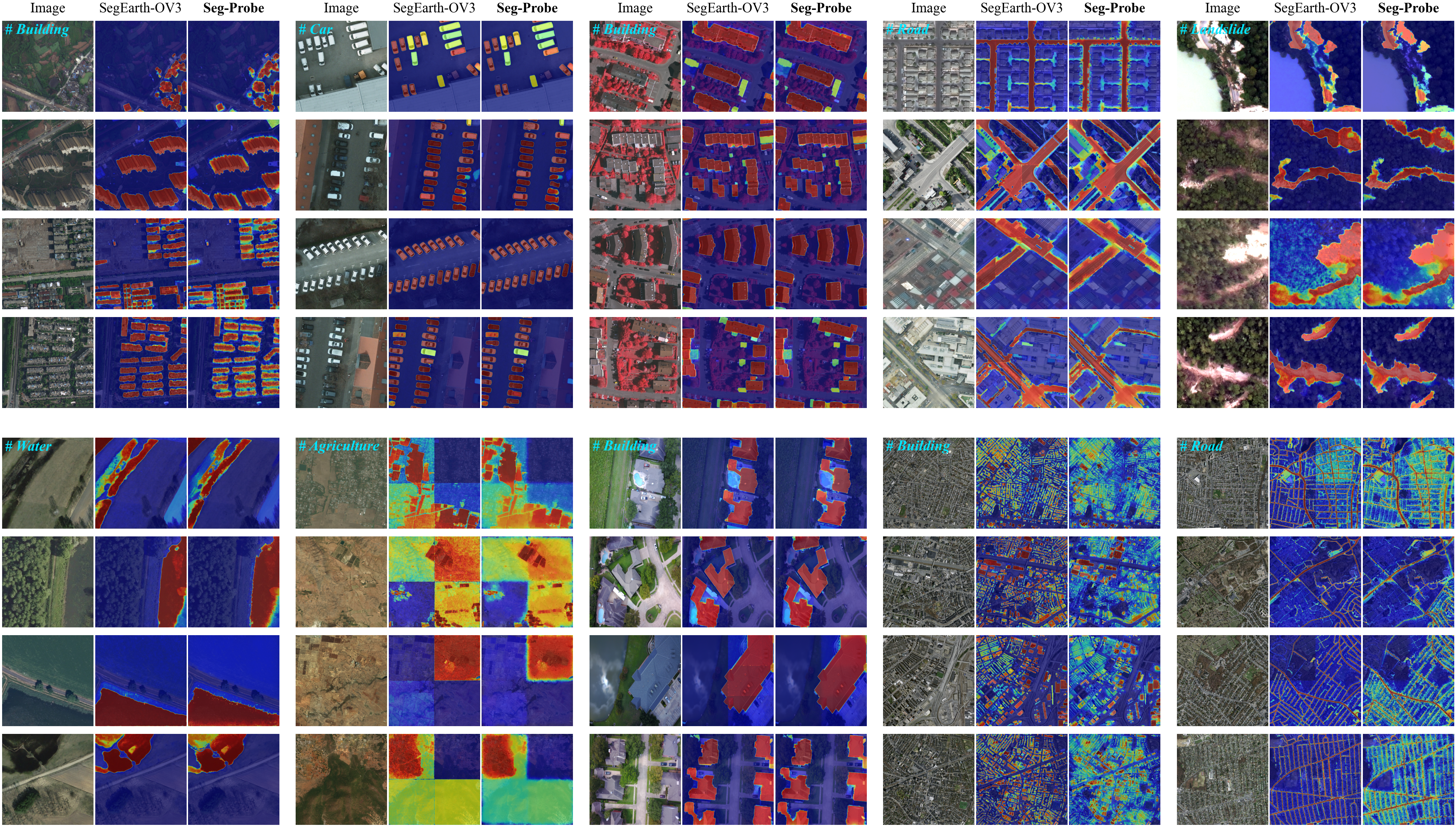}
\caption{\textbf{Class-evidence heatmaps of Seg-Probe and its SegEarth-OV3
backbone on the ten audited datasets.}
Every panel follows the same column protocol---Image / SegEarth-OV3 /
Seg-Probe---and the cyan \texttt{\# Class} tag names the queried class.
Relative to the backbone, Seg-Probe concentrates evidence on the probed class
and suppresses background bleed, so its disagreements with the human masks are
worth adjudicating rather than dismissing as backbone noise.}
\label{fig:heatmap}
\end{figure*}

\subsection{Are the Candidate Masks Strong Enough?}
\label{sec:segresults}

\begin{figure}[t]
\centering
\includegraphics[width=\linewidth]{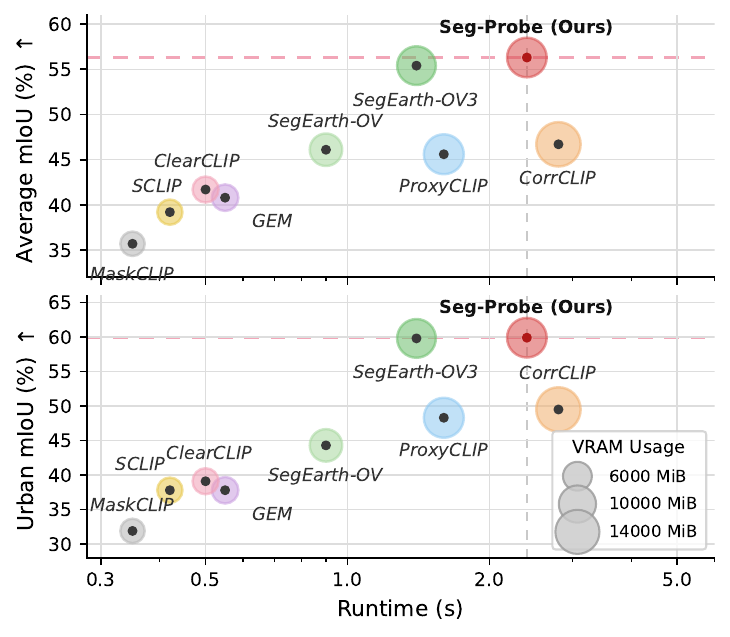}
\caption{\textbf{Accuracy--efficiency trade-off of training-free
open-vocabulary methods.} Average mIoU over the ten audited datasets
(\emph{top}) and fine-GSD urban mIoU, the mean of Potsdam and Vaihingen
(\emph{bottom}), against per-image runtime (log scale); bubble area encodes
peak inference VRAM. Dashed lines mark Seg-Probe, which attains the best
accuracy at a cost comparable to its backbone.}
\label{fig:scatter}
\end{figure}

A meaningful audit requires candidate masks whose disagreements with human
labels cannot be dismissed as poor predictions (RQ1).
\tabref{tab:model} shows that Seg-Probe matches or exceeds SegEarth-OV3 on
nine of ten datasets and achieves the best average performance
($56.3$ vs.\ $55.4$) among nine training-free methods.
\figref{fig:heatmap} shows where this margin comes from: across all ten
datasets, Seg-Probe concentrates its class evidence on the probed class and
suppresses it elsewhere, whereas the backbone bleeds onto surrounding context.
Its modules provide small but consistent gains, while
\figref{fig:scatter} shows that this improvement incurs only
$1.0$\,s additional inference time per image and negligible extra peak
VRAM.
Together, these results establish Seg-Probe as a strong and efficient
candidate generator, making its disagreements with human annotations
meaningful audit cases.
Additional ablations and efficiency analyses are provided in the
supplementary material.

\begin{table*}[t]
\centering
\small
\setlength{\tabcolsep}{4.2pt}
\renewcommand{\arraystretch}{1.12}
\begin{tabular}{@{}l l c c r ccc cc c c@{}}
\toprule
 & & & & &
\multicolumn{3}{c}{\textbf{Validity controls} (det.\ rate \%, $\uparrow$)} &
\multicolumn{2}{c}{\textbf{Mean fidelity}} & & \\
\cmidrule(lr){6-8}\cmidrule(lr){9-10}
\multirow{-2}{*}{\textbf{Dataset}} & \multirow{-2}{*}{\textbf{Platform}} &
\multirow{-2}{*}{\textbf{GSD\,(m)}} & \multirow{-2}{*}{\textbf{\#cls}} &
\multirow{-2}{*}{\textbf{\#pairs}} &
Dilate & Erode & Shift &
$\bar F_{\mathrm{gt}}$ & $\bar F_{\mathrm{pred}}$ &
\multirow{-2}{*}{$\boldsymbol{\Delta F}$} & \multirow{-2}{*}{\textbf{Pred-win (\%)}} \\
\midrule
\rowcolor{bandcol}
\multicolumn{12}{c}{\emph{Core five-dataset suite}} \\
Landslide       & UAV       & 0.20      & 2 & 5155 & 60 & 59 & 80 & $+0.029$ & $+0.039$ & $+0.010$ & \textbf{60.0} \\
LoveDA          & Satellite & 0.30      & 7 & 1070 & 34 & 70 & 84 & $+0.166$ & $+0.172$ & $+0.006$ & 49.9 \\
Potsdam         & Airborne  & 0.05      & 6 & 1128 & 61 & 58 & 89 & $+0.195$ & $+0.189$ & $-0.007$ & \textbf{50.4} \\
Vaihingen       & Airborne  & 0.09      & 6 &  469 & 57 & 64 & 92 & $+0.179$ & $+0.218$ & $+0.039$ & \textbf{71.6} \\
OpenEarthMap    & Mixed     & 0.25--0.5 & 9 &  828 & 53 & 80 & 80 & $+0.121$ & $+0.116$ & $-0.005$ & 49.0 \\
\midrule
\rowcolor{bandcol}
\multicolumn{12}{c}{\emph{Five datasets with documented annotation defects}} \\
LandCover.ai    & Airborne  & 0.25--0.5 & 5 &  497 & 35 & 83 & 89 & $+0.295$ & $+0.322$ & $+0.027$ & \textbf{56.5} \\
DeepGlobe       & Satellite & 0.50      & 7 &  572 & 56 & 52 & 78 & $+0.097$ & $+0.077$ & $-0.020$ & 43.9 \\
FloodNet        & UAV       & var.      &10 &  752 & 60 & 61 & 80 & $+0.127$ & $+0.175$ & $+0.048$ & \textbf{57.4} \\
Mass.\ Buildings & Airborne  & 1.00      & 2 &  130 &  0 & 80 & 70 & $-0.284$ & $-0.280$ & $+0.003$ & \textbf{55.4} \\
Mass.\ Roads     & Airborne  & 1.00      & 2 &  130 & 15 & 86 & 77 & $-0.038$ & $+0.005$ & $+0.044$ & \textbf{76.2} \\
\midrule
\textbf{All}     & ---       & 0.05--1   & --& \textbf{10731} & \textbf{54} & \textbf{63} & \textbf{82} & $\boldsymbol{+0.092}$ & $\boldsymbol{+0.102}$ & $\boldsymbol{+0.010}$ & \textbf{56.6} \\
\bottomrule
\end{tabular}
\caption{\textbf{Benchmark statistics and the CMF audit over $10{,}731$
image-class pairs.}
\emph{Validity controls}: rate at which CMF ranks the clean GT mask above a
corrupted version. $\Delta F=\bar F_{\mathrm{pred}}-\bar F_{\mathrm{gt}}$;
\emph{Pred-win}: fraction of pairs with $\Delta F>0$ (bold when the
prediction wins the majority). The ``All'' row pools all pairs.}
\label{tab:main}
\end{table*}

\subsection{Does CMF Measure Mask Fidelity?}
\label{sec:validity}

We first test whether CMF responds correctly to controlled mask corruptions
(RQ2). As shown in \tabref{tab:main}, CMF detects translation on
$70$--$92\%$ of image--class pairs and erosion on $52$--$86\%$, but is less
sensitive to dilation ($0$--$61\%$). This pattern indicates that CMF
primarily captures semantic displacement and omission rather than minor
boundary expansion. Moreover, its contrastive design reduces the correlation
with mask area from $\rho{=}0.61$ to $\rho{=}0.08$. Because these controls
still assume that the original annotation is clean, we provide an independent
expert-grounded validation in \secref{sec:human}.

\begin{figure*}[t]
\centering
\includegraphics[width=\textwidth]{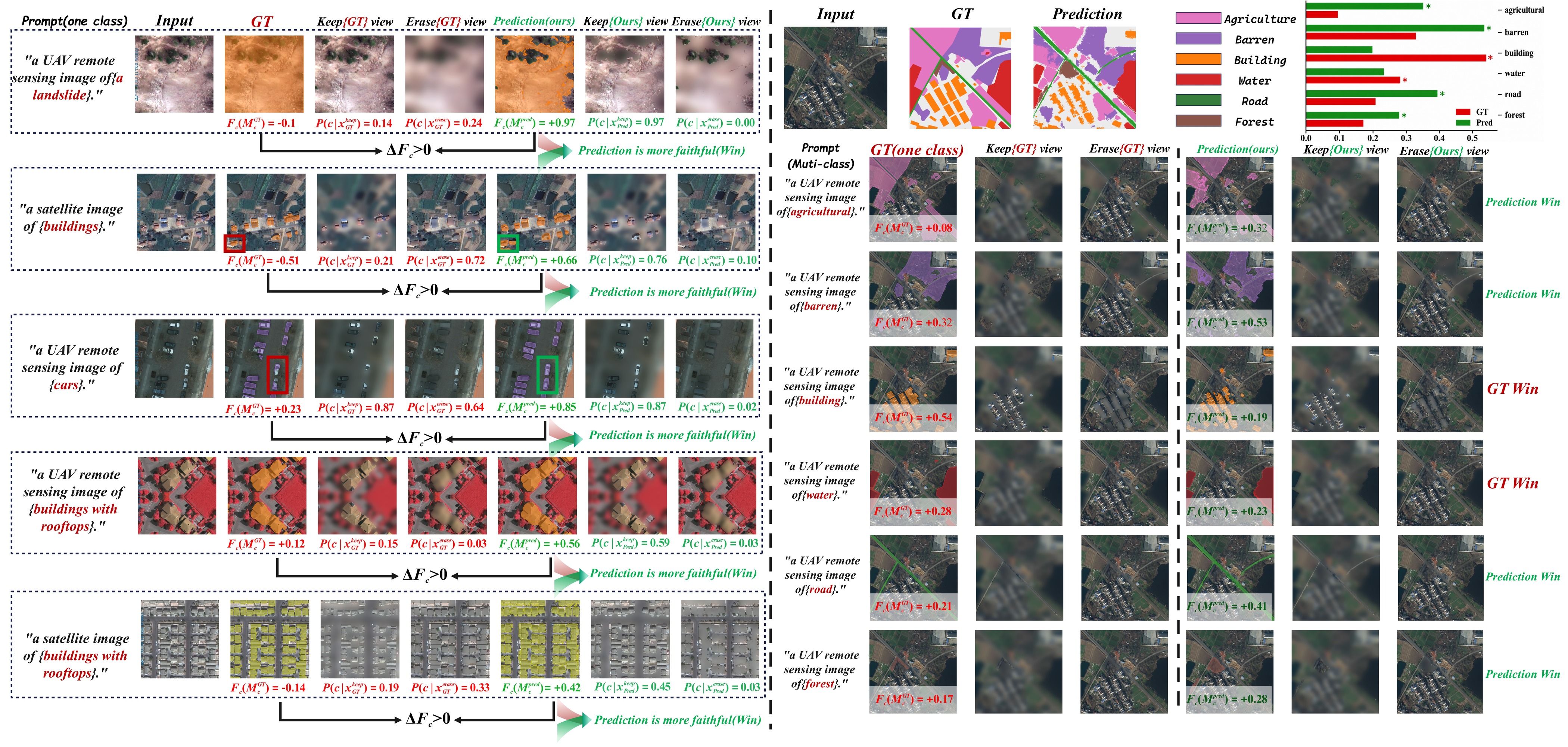}
\caption{\textbf{The CMF judging process.}
\emph{Left}: single-class audits on five datasets---the judge scores the
keep/erase views of the ground truth and of the prediction, and the
fidelity gap $\Delta F_c$ decides the verdict.
\emph{Right}: multi-class audit of one LoveDA image---each annotated class
is audited independently; CMF prefers the prediction on four classes but
defers to the ground truth on \emph{building} and \emph{water}, so the
audit is class-wise and not model-biased.}
\label{fig:evidence}
\end{figure*}

\subsection{What Does the CMF Audit Reveal?}
\label{sec:audit}

We next apply CMF to the $10{,}731$ audited image--class pairs (RQ3).
\tabref{tab:main} shows that predictions are preferred on seven of ten
datasets, with a pooled win rate of $56.6\%$ and a dataset-macro average of
$57.0\%$. More importantly, the verdicts exhibit a consistent class
structure across resolutions and platforms: well-defined objects such as
buildings, cars, and roads tend to favor predictions, whereas amorphous
land-cover classes such as cropland, agriculture, and urban areas more often
favor human annotations.
The gap is large enough to be class-diagnostic rather than marginal: cars
reach $84.6\%$ and buildings $75.2\%$ on Vaihingen, buildings
$76.6$--$77.9\%$ on LandCover.ai and FloodNet, and roads $76.2\%$ on
Massachusetts Roads, whereas LoveDA agriculture falls to $35.4\%$ and
DeepGlobe urban land to $23.5\%$.
The split tracks how annotations are produced rather than how difficult a
class is to see: compact man-made objects have crisp, verifiable outlines that
polygonization and minimum-mapping-unit conventions systematically simplify,
whereas land-cover boundaries are partly a matter of regional convention, so
the human label carries information the image alone does not, and CMF defers
to it.
\figref{fig:evidence} makes the underlying evidence explicit and shows
why a verdict is auditable rather than opaque.
On the left, five single-class cases print the entire basis of the decision:
the image, each mask with its keep and erase composites, and the judge's
class probability on every view, so the sign of $\Delta F_c$ can be read off
directly and no reference mask enters the computation.
On the right, one LoveDA image is audited class by class: CMF prefers the
prediction on \emph{agricultural}, \emph{barren}, \emph{road}, and
\emph{forest}, yet defers to the ground truth on \emph{building} and
\emph{water}, where the annotation already delineates the class well.
A single image can therefore split its verdicts across classes, which is what
distinguishes a class-wise audit from a global preference for the stronger
model.
These win rates are pair-level statistics conditioned on each dataset's
annotated classes and vocabulary. Full class-wise results and further
judging walkthroughs are provided in the supplementary material.

\subsection{Expert-Grounded Validation of CMF}
\label{sec:human}
To independently validate CMF, we collect expert judgments on $300$
image--class pairs from six representative datasets. Three remote-sensing
annotators compare the image with two anonymized and randomly ordered masks,
and their majority vote defines the expert consensus.
Twenty-one pairs received a split vote and are excluded, leaving $279$
decidable pairs, and the three reference rules are keep-only scoring, which
discards the erase term, SAM\,3's own presence confidence, and a trained
label-quality model. As shown in
\subtabref{tab:panel}{a}, CMF achieves $81.0\%$ agreement with the
consensus, outperforming simpler adjudication rules by at least $10.3$
percentage points, and the bootstrap intervals do not overlap
($[76.3,85.3]$ against $[65.4,75.8]$ for the strongest baseline).
The advantage is widest precisely where the audit reports the strongest class
structure, reaching $89.3\%$ on man-made classes, and narrows to $70.5\%$ on
amorphous land cover, so CMF is most reliable on the classes for which it most
often overturns the annotation. Agreement further increases to $88.4\%$ for
high-margin verdicts, indicating that $|\Delta F|$ reflects decision
reliability
(\subtabref{tab:panel}{c}). Full protocols, baseline definitions, and detailed
results are provided in the supplementary material.

\begin{table*}[t]
\centering
\scriptsize
\setlength{\tabcolsep}{2.2pt}
\renewcommand{\arraystretch}{1.08}
\begin{minipage}[t]{0.485\textwidth}
\centering
{\footnotesize\textbf{(a)} Expert adjudication on $279$ non-tie pairs
(from $300$; $21$ majority ties excluded).
Blue: gain vs.\ strongest baseline (\underline{underlined}).}\\[2pt]
\begin{tabular}{@{}l ccccc@{}}
\toprule
\textbf{Method} & \textbf{Overall} & \textbf{Bal.} & \textbf{High-m.} & \textbf{Man-m.} & \textbf{Amorph.} \\
\midrule
Keep-only         & 64.3 & 63.1 & 69.5 & 67.9 & 59.8 \\
SAM\,3 confidence & 68.0 & 66.5 & 72.3 & 71.4 & 63.6 \\
Label-quality     & \underline{70.7} & \underline{69.2} & \underline{74.5} & \underline{73.2} & \underline{67.4} \\
\midrule
\rowcolor{bandcol}
\textbf{CMF (ours)}
& \cmfcell{81.0}{10.3}
& \cmfcell{79.6}{10.4}
& \cmfcell{88.4}{13.9}
& \cmfcell{89.3}{16.1}
& \cmfcell{70.5}{3.1} \\
\bottomrule
\end{tabular}
\end{minipage}%
\hfill
\begin{minipage}[t]{0.485\textwidth}
\centering
{\footnotesize\textbf{(b)} Cross-domain mIoU (\%; mean, 3 seeds; std\,$\leq$\,0.6).
$\Delta$: vs.\ strongest control (\underline{underlined}).}\\[1pt]
\begin{tabular}{@{}l ccccc c@{}}
\toprule
\textbf{Transfer} & \textbf{Raw} & \textbf{Rand.} & \textbf{Seg-P.} & \textbf{Conf.} & \textbf{CMF} & $\boldsymbol{\Delta}$ \\
\midrule
LoveDA$\rightarrow$OEM       & 41.6 & 41.9 & 43.0 & \underline{43.9} & \textbf{45.5} & \dgain{1.6} \\
OEM$\rightarrow$LoveDA       & 43.8 & 44.1 & 45.0 & \underline{45.9} & \textbf{47.1} & \dgain{1.2} \\
Potsdam$\rightarrow$Vaihingen& 47.9 & 48.1 & 48.6 & \underline{49.5} & \textbf{50.8} & \dgain{1.3} \\
Vaihingen$\rightarrow$Potsdam& 49.6 & 49.8 & 50.1 & \underline{50.9} & \textbf{52.2} & \dgain{1.3} \\
\midrule
\rowcolor{bandcol}
\textbf{Avg.}                & 45.7 & 46.0 & 46.7 & \underline{47.6} & \textbf{48.8} & \dgain{1.2} \\
\bottomrule
\end{tabular}
\end{minipage}

\vspace{7pt}

\begin{minipage}[t]{0.485\textwidth}
\centering
{\footnotesize\textbf{(c)} Arbitration margin $\delta$: coverage versus
precision.}\\[2pt]
\setlength{\tabcolsep}{3.2pt}
\begin{tabular}{@{}cccc@{}}
\toprule
$\boldsymbol{\delta}$ & \textbf{Cov.} & \textbf{Agr.} & \textbf{mIoU} \\
\midrule
\rowcolor{bandcol}
\textbf{0.00} & \textbf{56.6} & 81.0 & \textbf{48.8} \\
0.05 & 41.2 & 85.7 & 48.7 \\
0.10 & 30.7 & 88.9 & 48.5 \\
0.15 & 22.4 & 90.3 & 48.0 \\
\bottomrule
\end{tabular}
\end{minipage}%
\hfill
\begin{minipage}[t]{0.485\textwidth}
\centering
{\footnotesize\textbf{(d)} Judge swap: stability of the audit under four
frozen judges.}\\[2pt]
\setlength{\tabcolsep}{3.2pt}
\begin{tabular}{@{}lccc@{}}
\toprule
\textbf{Judge} & \textbf{Bldg} & $\boldsymbol{\rho}$ & \textbf{Dom.} \\
\midrule
CLIP                & 64.8 & 0.81 & gen. \\
OpenCLIP            & 68.2 & 0.86 & gen. \\
GeoRSCLIP           & 70.9 & 0.92 & RS \\
\rowcolor{bandcol}
\textbf{RemoteCLIP} & \textbf{72.7} & \textbf{1.00} & RS \\
\bottomrule
\end{tabular}
\end{minipage}
\caption{\textbf{Expert validation, transfer, and sensitivity.}
(a)~Adjudication vs.\ expert consensus (\secref{sec:human}); CMF CI
$[76.3,85.3]$ vs.\ label-quality $[65.4,75.8]$.
(b)~Shared-class transfer; Raw/Rand./Seg-P./Conf.\ = annotation / random /
Seg-Probe-only / coverage-matched confidence.
(c)~Cov./Agr.\ = replacement coverage / expert agreement; mIoU = avg.\ of~(b);
default $\delta{=}0$.
(d)~Bldg = building pred-win; $\rho$ = Spearman vs.\ RemoteCLIP; Dom.\ =
pretraining domain.}
\label{tab:panel}
\end{table*}

\subsection{Cross-Domain Validation}
\label{sec:crossdomain}

If CMF verdicts carry real signal, the supervision they select should be more
useful than the raw annotations (RQ4).
Two questions must be separated here.
First, does arbitrated supervision transfer better?
Second---and more subtle---if it does, is the gain attributable to CMF's
\emph{selection}, or merely to injecting strong foundation-model
pseudo-labels into the training set?
We answer both with matched controls.

\begin{figure*}[t]
\centering
\includegraphics[width=\textwidth]{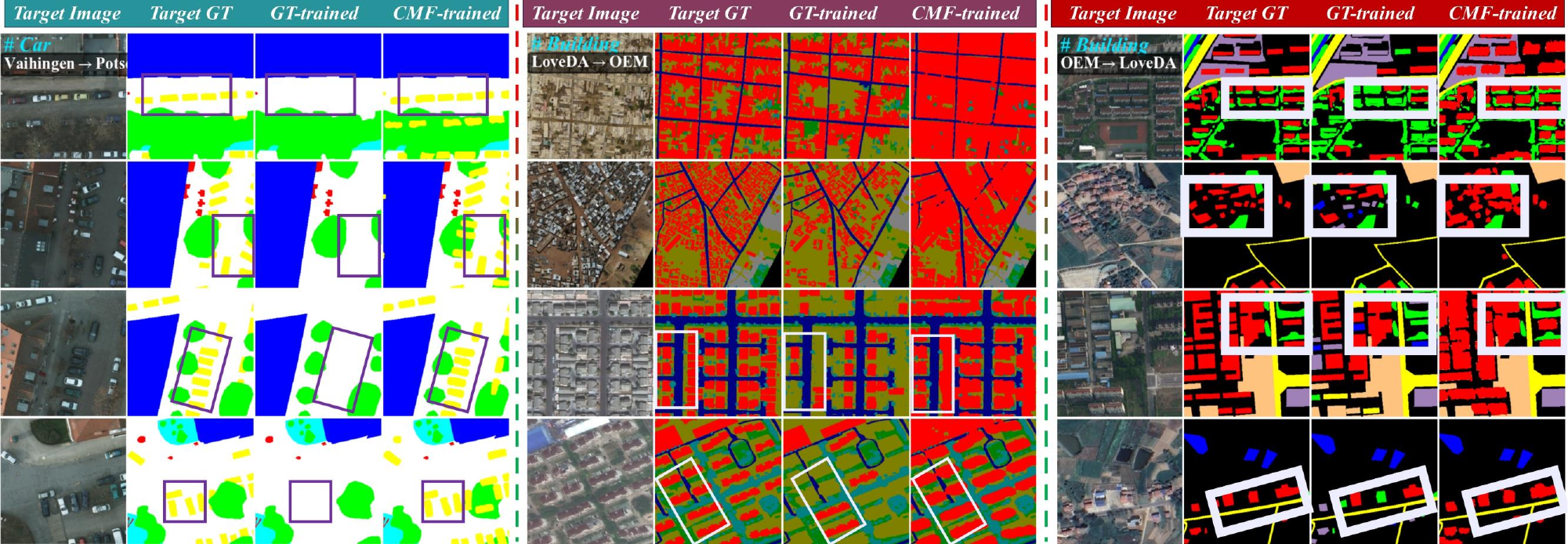}
\caption{\textbf{Cross-domain qualitative comparison on three target
domains.} On unseen target images, the model trained on CMF-arbitrated
labels recovers buildings and roads that the GT-trained model misses
(boxes), consistent with the audit's class structure.}
\label{fig:crossviz}
\end{figure*}

\paragraph{Protocol.}
We construct the arbitrated map $\tilde Y$ via \equref{eq:arbitration}
and train the same DeepLabV3+~\citep{deeplabv3plus} segmenter on a
\emph{source} dataset under five supervisions:
(i)~\textbf{raw Annotation} $Y^a$;
(ii)~\textbf{Seg-Probe only}, i.e., pure predictions $Y^p$;
(iii)~\textbf{confidence replacement}, which substitutes the prediction
whenever SAM\,3's presence confidence exceeds a threshold calibrated to match
CMF's replacement coverage;
(iv)~\textbf{random replacement}, which substitutes a random subset of pairs
matched to CMF in pair count, class ratio, and replaced-pixel fraction; and
(v)~\textbf{CMF replacement} $\tilde Y$.
Architecture, optimizer, and schedule are identical; each configuration is
trained with three random seeds.
Models are evaluated \emph{without fine-tuning} on a \emph{target} dataset
that shares the audited classes but differs in geography, sensor, or
resolution; we report the four bidirectional multi-class directions in the
main text and defer class-level transfers (building, road, car; gains of
$+4.8$ to $+6.6$ IoU under the same protocol) to the supplementary material.

\paragraph{Results.}
\subtabref{tab:panel}{b} shows a consistent ordering on all four
directions:
\emph{CMF $>$ confidence $>$ Seg-Probe-only $>$ raw GT}, with random
replacement statistically indistinguishable from raw GT.
Three observations disentangle the effect.
First, CMF beats \emph{Seg-Probe only} by $+2.1$ mIoU on average, so the
gain is not pure pseudo-label distillation---indiscriminately trusting the
stronger model is worse than selecting where to trust it.
Second, CMF beats \emph{confidence replacement} by $+1.2$ at matched
coverage, so the image-conditioned keep--erase contrast selects better than
the segmenter's own confidence.
Third, \emph{random replacement} at matched coverage yields no benefit
($+0.3$, within one std), ruling out regularization-by-label-mixing as the
explanation.
\figref{fig:crossviz} visualizes the effect on three target domains.
The gains are concentrated exactly where the audit predicted them: the
CMF-trained model closes building footprints and reconnects road segments
that the GT-trained model reproduces only partially, while amorphous land
cover---the classes on which CMF usually defers to the annotation---is
reproduced comparably by both models.
The qualitative pattern therefore mirrors the class structure of
\secref{sec:audit} rather than a uniform improvement, which is what one
expects if the transfer gain originates in selective arbitration.
We note the boundary of this claim: because the target annotations may
themselves be imperfect, transfer gains establish that CMF-selected
supervision is more \emph{useful} under the evaluated settings, not by
themselves that every selected mask is more faithful.

\paragraph{Arbitration-margin robustness.}
\subtabref{tab:panel}{c} sweeps the margin $\delta$ of
\equref{eq:arbitration}.
Raising $\delta$ shrinks replacement coverage while increasing expert
agreement on the replaced pairs, yet transfer mIoU varies by at most
$0.8$ across the sweep and is highest at the default $\delta{=}0$: the
transfer gain is not driven by a tuned threshold, and low-margin
replacements are, on aggregate, harmless.
Practitioners who require higher per-replacement precision can trade
coverage for safety at $\delta{>}0$ without losing most of the gain.

\subsection{Ablations and Sensitivity}
\label{sec:ablation}

Compositing, prompting, and temperature choices are stable: Gaussian-blur
fill outperforms black and mean-color fills (shift detection $0.85$ vs.\
$0.71$/$0.79$), synonym ensembling edges out a single template, and verdicts
vary by at most $0.6$ points of win rate across
$\tau\in\{50,100,200\}$; the full design sweep is in the supplementary
material.
The critical sensitivity is the judge itself, so \subtabref{tab:panel}{d}
swaps RemoteCLIP for CLIP~\citep{clip}, OpenCLIP~\citep{openclip}, and
GeoRSCLIP~\citep{georsclip}: all alternatives reproduce the man-made vs.\
amorphous class structure ($\rho{\geq}0.8$ agreement between per-class
win-rate profiles), so the qualitative audit conclusion does not hinge on the
specific judge, although all four judges share the CLIP training paradigm and
may share some biases.

\subsection{Threats to Validity and Limitations}
\label{sec:threats}
Three considerations delimit our conclusions.
\emph{Judge and metric scope.}
CMF inherits the biases of its frozen judge and measures class-relative
\emph{semantic} fidelity under a specified vocabulary rather than precise
boundary geometry or instance topology, so a mask that is semantically right
but geometrically loose can win a comparison that a boundary metric would
decide differently.
Using distinct models for generation and adjudication reduces, but does not
eliminate, correlated failures, since all four judges we test share the CLIP
pretraining paradigm.
\emph{Audit and validation scope.}
The audit is label-conditioned: it excludes classes that only the prediction
claims and it is defined relative to each dataset's vocabulary, so win rates
should be read as pair-level statistics rather than absolute annotation error
rates.
Cross-domain gains likewise establish that CMF-selected supervision is more
useful under the evaluated transfers, not that every selected mask is more
faithful, because the target annotations are themselves imperfect.
\emph{Expert-study scale.}
The blinded study covers $300$ pairs from six datasets and three annotators,
which supports the aggregate verdict trends but leaves broader panels to
future work---especially for ambiguous land cover, the regime in which CMF
agrees with the expert consensus least often.

\subsection{Implementation and Supplementary Evidence}
\label{sec:implementation}
Three aspects support the reproducibility, scalability, and transparency of
our study.
\emph{(1) Implementation and efficiency.}
Stages~I--II are training-free: Seg-Probe uses a frozen SAM\,3 encoder,
whereas CMF employs a frozen RemoteCLIP ViT-L/14 judge. Unless otherwise
specified, judge inputs are resized to $224\times224$, keep/erase views use
Gaussian blur with $\sigma=24$, and the logit scale is fixed to $\tau=100$.
All experiments are conducted on a single NVIDIA GeForce RTX~4090 GPU with
24\,GB memory. After candidate generation, CMF requires four judge forward
passes per image--class pair and audits all $10{,}731$ pairs in approximately
16 minutes.
\emph{(2) Supplementary validation.}
To keep the main paper focused on the central evidence chain, the
supplementary material provides the complete Seg-Probe formulation and
ablations, CMF implementation and design analyses, corruption-response and
area-sensitivity experiments, and full dataset- and class-level audit results
with confidence intervals and statistical tests. It also includes extensive
keep/erase evidence walkthroughs, disagreement atlases, and representative
success and failure cases.
\emph{(3) Reproducibility and responsible use.}
The supplementary material further documents the expert-study protocol and
adjudication baselines, cross-domain training settings, matched-control
construction, class-level transfer and arbitration-budget analyses, together
with computational resources, ethical considerations, and intended-use
guidance.

\section{Conclusion}
\label{sec:conclusion}
We asked whether a mask disagreement can be adjudicated without treating the
human annotation as an oracle, and answered with CMF: a reference-free
keep--erase contrast scored by a frozen external vision-language judge.
On a blinded expert consensus, CMF adjudicates substantially better than
keep-only scoring, model confidence, and a trained label-quality baseline;
applied to $10{,}731$ image-class pairs across ten benchmarks, it reveals
systematic, class-structured annotation distortion; and its
conservative class-wise arbitration improves cross-domain transfer over raw
labels, pure predictions, and matched replacement controls, showing the
gain comes from selection rather than pseudo-label distillation.
CMF is a scalable screening tool, not a replacement for expert annotation:
it surfaces disagreements that overlap metrics cannot resolve and
prioritizes masks for human review.

Two directions follow naturally.
First, CMF scores a region claim against a class name, so the same
keep--erase contrast applies wherever such a claim can be phrased as text;
extending it to instance and panoptic claims, or to open-vocabulary classes
that a schema never annotated, would widen the audit beyond the
label-conditioned setting used here.
Second, the arbitration margin behaves like a precision--coverage dial, which
suggests deploying CMF as a triage layer in annotation pipelines:
high-margin disagreements can be corrected automatically, mid-margin ones
routed to human review, and low-margin ones left untouched.
More broadly, we hope this audit reframes a habit rather than adds a
benchmark. Once masks can be compared against the image instead of against
their overlap with a reference, asking how much of a benchmark's remaining
headroom is annotation noise becomes a measurement one can report rather than
an assumption one has to make.

\section*{Acknowledgments}
We thank the three remote-sensing annotators who took part in the blinded
expert study for their careful and independent judgments, and the maintainers
of the ten public benchmarks audited here for releasing the imagery and
annotations that made this analysis possible.

\clearpage
\bibliography{references}

@inproceedings{sam3,
  title        = {{SAM} 3: Segment Anything with Concepts},
  author       = {Carion, Nicolas and Gustafson, Laura and Hu, Yuan-Ting and
                  Debnath, Shoubhik and Hu, Ronghang and
                  Suris Coll-Vinent, Didac and Ryali, Chaitanya and
                  Alwala, Kalyan Vasudev and Khedr, Haitham and
                  Huang, Andrew and Lei, Jie and Ma, Tengyu and
                  Guo, Baishan and Kalla, Arpit and Marks, Markus and
                  Greer, Joseph and Wang, Meng and Sun, Peize and
                  R{\"a}dle, Roman and Afouras, Triantafyllos and
                  Mavroudi, Effrosyni and Xu, Katherine and Wu, Tsung-Han and
                  Zhou, Yu and Momeni, Liliane and Hazra, Rishi and
                  Ding, Shuangrui and Vaze, Sagar and Porcher, Francois and
                  Li, Feng and Li, Siyuan and Kamath, Aishwarya and
                  Cheng, Ho Kei and Doll{\'a}r, Piotr and Ravi, Nikhila and
                  Saenko, Kate and Zhang, Pengchuan and
                  Feichtenhofer, Christoph},
  booktitle    = {International Conference on Learning Representations (ICLR)},
  year         = {2026},
  url          = {https://openreview.net/forum?id=r35clVtGzw},
  eprint       = {2511.16719},
  archivePrefix= {arXiv},
  primaryClass = {cs.CV}
}

@inproceedings{corrclip,
  title        = {{CorrCLIP}: Reconstructing Patch Correlations in {CLIP} for
                  Open-Vocabulary Semantic Segmentation},
  author       = {Zhang, Dengke and Liu, Fagui and Tang, Quan},
  booktitle    = {Proceedings of the IEEE/CVF International Conference on
                  Computer Vision (ICCV)},
  pages        = {24677--24687},
  month        = oct,
  year         = {2025}
}

@inproceedings{sam,
  title        = {Segment Anything},
  author       = {Kirillov, Alexander and Mintun, Eric and Ravi, Nikhila and
                  Mao, Hanzi and Rolland, Chloe and Gustafson, Laura and
                  Xiao, Tete and Whitehead, Spencer and Berg, Alexander C. and
                  Lo, Wan-Yen and Doll{\'a}r, Piotr and Girshick, Ross},
  booktitle    = {Proceedings of the IEEE/CVF International Conference on
                  Computer Vision (ICCV)},
  pages        = {4015--4026},
  month        = oct,
  year         = {2023}
}

@inproceedings{sam2,
  title        = {{SAM} 2: Segment Anything in Images and Videos},
  author       = {Ravi, Nikhila and Gabeur, Valentin and Hu, Yuan-Ting and
                  Hu, Ronghang and Ryali, Chaitanya and Ma, Tengyu and
                  Khedr, Haitham and R{\"a}dle, Roman and Rolland, Chloe and
                  Gustafson, Laura and Mintun, Eric and Pan, Junting and
                  Alwala, Kalyan Vasudev and Carion, Nicolas and Wu, Chao-Yuan and
                  Girshick, Ross and Doll{\'a}r, Piotr and Feichtenhofer, Christoph},
  booktitle    = {International Conference on Learning Representations (ICLR)},
  year         = {2025},
  url          = {https://openreview.net/forum?id=Ha6RTeWMd0}
}

@inproceedings{clip,
  title        = {Learning Transferable Visual Models From Natural Language
                  Supervision},
  author       = {Radford, Alec and Kim, Jong Wook and Hallacy, Chris and
                  Ramesh, Aditya and Goh, Gabriel and Agarwal, Sandhini and
                  Sastry, Girish and Askell, Amanda and Mishkin, Pamela and
                  Clark, Jack and Krueger, Gretchen and Sutskever, Ilya},
  booktitle    = {Proceedings of the 38th International Conference on Machine
                  Learning},
  series       = {Proceedings of Machine Learning Research},
  volume       = {139},
  pages        = {8748--8763},
  publisher    = {PMLR},
  year         = {2021}
}

@article{georsclip,
  title        = {{RS5M} and {GeoRSCLIP}: A Large-Scale Vision-Language Dataset
                  and a Large Vision-Language Model for Remote Sensing},
  author       = {Zhang, Zilun and Zhao, Tiancheng and Guo, Yulong and Yin, Jianwei},
  journal      = {IEEE Transactions on Geoscience and Remote Sensing},
  volume       = {62},
  pages        = {1--23},
  year         = {2024},
  note         = {Art. no. 5642123},
  doi          = {10.1109/TGRS.2024.3449154}
}

@article{remoteclip,
  title        = {{RemoteCLIP}: A Vision Language Foundation Model for Remote
                  Sensing},
  author       = {Liu, Fan and Chen, Delong and Guan, Zhangqingyun and
                  Zhou, Xiaocong and Zhu, Jiale and Ye, Qiaolin and
                  Fu, Liyong and Zhou, Jun},
  journal      = {IEEE Transactions on Geoscience and Remote Sensing},
  volume       = {62},
  pages        = {1--16},
  year         = {2024},
  note         = {Art. no. 5622216},
  doi          = {10.1109/TGRS.2024.3390838}
}

@inproceedings{clipscore,
  title        = {{CLIPScore}: A Reference-free Evaluation Metric for Image
                  Captioning},
  author       = {Hessel, Jack and Holtzman, Ari and Forbes, Maxwell and
                  Le Bras, Ronan and Choi, Yejin},
  booktitle    = {Proceedings of the 2021 Conference on Empirical Methods in
                  Natural Language Processing},
  pages        = {7514--7528},
  publisher    = {Association for Computational Linguistics},
  address      = {Online and Punta Cana, Dominican Republic},
  month        = nov,
  year         = {2021},
  doi          = {10.18653/v1/2021.emnlp-main.595}
}

@inproceedings{segearthov,
  title        = {{SegEarth-OV}: Towards Training-Free Open-Vocabulary
                  Segmentation for Remote Sensing Images},
  author       = {Li, Kaiyu and Liu, Ruixun and Cao, Xiangyong and Bai, Xueru and
                  Zhou, Feng and Meng, Deyu and Wang, Zhi},
  booktitle    = {Proceedings of the IEEE/CVF Conference on Computer Vision and
                  Pattern Recognition (CVPR)},
  pages        = {10545--10556},
  month        = jun,
  year         = {2025}
}

@article{segearthov3,
  title        = {{SegEarth-OV3}: Exploring {SAM} 3 for Open-Vocabulary Semantic
                  Segmentation in Remote Sensing Images},
  author       = {Li, Kaiyu and Zhang, Shengqi and Wang, Yujie and Deng, Yupeng and
                  Wang, Zhi and Meng, Deyu and Cao, Xiangyong},
  journal      = {arXiv preprint arXiv:2512.08730},
  year         = {2025},
  eprint       = {2512.08730},
  archivePrefix= {arXiv},
  primaryClass = {cs.CV}
}

@inproceedings{reattnclip,
  title        = {{ReAttnCLIP}: Training-Free Open-Vocabulary Remote Sensing
                  Image Segmentation via Re-defined Attention in {CLIP}},
  author       = {Niu, Xin and Zhao, Manqi and Jiang, Dongsheng and Wu, Yingying
                  and Su, Bing},
  booktitle    = {Proceedings of the IEEE/CVF Conference on Computer Vision and
                  Pattern Recognition (CVPR)},
  pages        = {24980--24989},
  month        = jun,
  year         = {2026}
}

@inproceedings{coninfer,
  title        = {{ConInfer}: Context-Aware Inference for Training-Free
                  Open-Vocabulary Remote Sensing Segmentation},
  author       = {Chen, Wenyang and Hu, Zhanxuan and Zhang, Yaping and Ning,
                  Hailong and Tai, Yonghang},
  booktitle    = {Proceedings of the IEEE/CVF Conference on Computer Vision and
                  Pattern Recognition (CVPR) Findings},
  pages        = {7408--7418},
  month        = jun,
  year         = {2026}
}

@article{ovrs,
  title        = {Open-Vocabulary High-Resolution Remote Sensing Image Semantic
                  Segmentation},
  author       = {Cao, Qinglong and Chen, Yuntian and Ma, Chao and Yang, Xiaokang},
  journal      = {IEEE Transactions on Geoscience and Remote Sensing},
  volume       = {63},
  pages        = {1--14},
  year         = {2025},
  note         = {Art. no. 5620614},
  doi          = {10.1109/TGRS.2025.3559557}
}

@inproceedings{gsnet,
  title        = {Towards Open-Vocabulary Remote Sensing Image Semantic
                  Segmentation},
  author       = {Ye, Chengyang and Zhuge, Yunzhi and Zhang, Pingping},
  booktitle    = {Proceedings of the AAAI Conference on Artificial Intelligence},
  volume       = {39},
  number       = {9},
  pages        = {9436--9444},
  year         = {2025},
  doi          = {10.1609/aaai.v39i9.33022}
}

@article{lad2023segquality,
  title        = {Estimating Label Quality and Errors in Semantic Segmentation
                  Data via Any Model},
  author       = {Lad, Vedang and Mueller, Jonas},
  journal      = {arXiv preprint arXiv:2307.05080},
  year         = {2023},
  eprint       = {2307.05080},
  archivePrefix= {arXiv},
  primaryClass = {cs.CV}
}

@article{aio2,
  title        = {{AIO2}: Online Correction of Object Labels for Deep Learning
                  With Incomplete Annotation in Remote Sensing Image
                  Segmentation},
  author       = {Liu, Chenying and Albrecht, Conrad M. and Wang, Yi and
                  Li, Qingyu and Zhu, Xiao Xiang},
  journal      = {IEEE Transactions on Geoscience and Remote Sensing},
  volume       = {62},
  pages        = {1--17},
  year         = {2024},
  doi          = {10.1109/TGRS.2024.3373908}
}

@article{segassess,
  title        = {{SegAssess}: Panoramic Quality Mapping for Robust and
                  Transferable Unsupervised Segmentation Assessment},
  author       = {Yang, Bingnan and Zhang, Mi and Zhang, Zhili and Zhang, Zhan
                  and Zhao, Yuanxin and Hu, Xiangyun and Gong, Jianya},
  journal      = {IEEE Transactions on Geoscience and Remote Sensing},
  volume       = {63},
  pages        = {1--26},
  year         = {2025},
  doi          = {10.1109/TGRS.2025.3641753}
}

@inproceedings{li2023noisyseg,
  title        = {Semi-Supervised Semantic Segmentation under Label Noise via
                  Diverse Learning Groups},
  author       = {Li, Peixia and Purkait, Pulak and Ajanthan, Thalaiyasingam and
                  Abdolshah, Majid and Garg, Ravi and Husain, Hisham and Xu,
                  Chenchen and Gould, Stephen and Ouyang, Wanli and van den
                  Hengel, Anton},
  booktitle    = {Proceedings of the IEEE/CVF International Conference on
                  Computer Vision (ICCV)},
  pages        = {1229--1238},
  month        = oct,
  year         = {2023}
}

@inproceedings{prometheusvision,
  title        = {{Prometheus-Vision}: Vision-Language Model as a Judge for
                  Fine-Grained Evaluation},
  author       = {Lee, Seongyun and Kim, Seungone and Park, Sue and Kim,
                  Geewook and Seo, Minjoon},
  booktitle    = {Findings of the Association for Computational Linguistics:
                  ACL 2024},
  pages        = {11286--11315},
  publisher    = {Association for Computational Linguistics},
  address      = {Bangkok, Thailand},
  month        = aug,
  year         = {2024},
  doi          = {10.18653/v1/2024.findings-acl.672}
}

@inproceedings{northcutt2021labelerrors,
  title        = {Pervasive Label Errors in Test Sets Destabilize Machine
                  Learning Benchmarks},
  author       = {Northcutt, Curtis G. and Athalye, Anish and Mueller, Jonas},
  booktitle    = {Proceedings of the Neural Information Processing Systems Track
                  on Datasets and Benchmarks},
  volume       = {1},
  year         = {2021}
}

@inproceedings{loveda,
  title        = {{LoveDA}: A Remote Sensing Land-Cover Dataset for Domain
                  Adaptive Semantic Segmentation},
  author       = {Wang, Junjue and Zheng, Zhuo and Ma, Ailong and Lu, Xiaoyan and
                  Zhong, Yanfei},
  booktitle    = {Proceedings of the Neural Information Processing Systems Track
                  on Datasets and Benchmarks},
  volume       = {1},
  year         = {2021}
}

@inproceedings{openearthmap,
  title        = {{OpenEarthMap}: A Benchmark Dataset for Global High-Resolution
                  Land Cover Mapping},
  author       = {Xia, Junshi and Yokoya, Naoto and Adriano, Bruno and
                  Broni-Bediako, Clifford},
  booktitle    = {Proceedings of the IEEE/CVF Winter Conference on Applications
                  of Computer Vision (WACV)},
  pages        = {6254--6264},
  month        = jan,
  year         = {2023}
}

@article{isprs2d,
  title        = {The {ISPRS} Benchmark on Urban Object Classification and
                  {3D} Building Reconstruction},
  author       = {Rottensteiner, Franz and Sohn, Gunho and Jung, Jaewook and
                  Gerke, Markus and Baillard, Caroline and Benitez, Sebastien and
                  Breitkopf, Uwe},
  journal      = {ISPRS Annals of the Photogrammetry, Remote Sensing and
                  Spatial Information Sciences},
  volume       = {I-3},
  pages        = {293--298},
  year         = {2012},
  doi          = {10.5194/isprsannals-I-3-293-2012}
}

@inproceedings{landcoverai,
  title        = {{LandCover.ai}: Dataset for Automatic Mapping of Buildings,
                  Woodlands, Water and Roads From Aerial Imagery},
  author       = {Boguszewski, Adrian and Batorski, Dominik and
                  Ziemba-Jankowska, Natalia and Dziedzic, Tomasz and
                  Zambrzycka, Anna},
  booktitle    = {Proceedings of the IEEE/CVF Conference on Computer Vision and
                  Pattern Recognition (CVPR) Workshops},
  pages        = {1102--1110},
  month        = jun,
  year         = {2021}
}

@inproceedings{deepglobe,
  title        = {{DeepGlobe} 2018: A Challenge to Parse the Earth Through
                  Satellite Images},
  author       = {Demir, Ilke and Koperski, Krzysztof and Lindenbaum, David and
                  Pang, Guan and Huang, Jing and Basu, Saikat and Hughes, Forest and
                  Tuia, Devis and Raskar, Ramesh},
  booktitle    = {Proceedings of the IEEE/CVF Conference on Computer Vision and
                  Pattern Recognition (CVPR) Workshops},
  pages        = {172--181},
  year         = {2018},
  doi          = {10.1109/CVPRW.2018.00031}
}

@article{floodnet,
  title        = {{FloodNet}: A High Resolution Aerial Imagery Dataset for Post
                  Flood Scene Understanding},
  author       = {Rahnemoonfar, Maryam and Chowdhury, Tashnim and
                  Sarkar, Argho and Varshney, Debvrat and Yari, Masoud and
                  Murphy, Robin Roberson},
  journal      = {IEEE Access},
  volume       = {9},
  pages        = {89644--89654},
  year         = {2021},
  doi          = {10.1109/ACCESS.2021.3090981}
}

@phdthesis{mnih2013,
  title        = {Machine Learning for Aerial Image Labeling},
  author       = {Mnih, Volodymyr},
  school       = {University of Toronto},
  year         = {2013}
}

@inproceedings{openclip,
  title        = {Reproducible Scaling Laws for Contrastive Language-Image
                  Learning},
  author       = {Cherti, Mehdi and Beaumont, Romain and Wightman, Ross and
                  Wortsman, Mitchell and Ilharco, Gabriel and Gordon, Cade and
                  Schuhmann, Christoph and Schmidt, Ludwig and Jitsev, Jenia},
  booktitle    = {Proceedings of the IEEE/CVF Conference on Computer Vision and
                  Pattern Recognition (CVPR)},
  pages        = {2818--2829},
  month        = jun,
  year         = {2023}
}

@inproceedings{deeplabv3plus,
  title        = {Encoder-Decoder with Atrous Separable Convolution for Semantic
                  Image Segmentation},
  author       = {Chen, Liang-Chieh and Zhu, Yukun and Papandreou, George and
                  Schroff, Florian and Adam, Hartwig},
  booktitle    = {Proceedings of the European Conference on Computer Vision
                  (ECCV)},
  pages        = {801--818},
  month        = sep,
  year         = {2018}
}

@inproceedings{maskclip,
  title        = {Extract Free Dense Labels from {CLIP}},
  author       = {Zhou, Chong and Loy, Chen Change and Dai, Bo},
  booktitle    = {Computer Vision -- ECCV 2022},
  series       = {Lecture Notes in Computer Science},
  volume       = {13688},
  pages        = {696--712},
  publisher    = {Springer},
  year         = {2022},
  doi          = {10.1007/978-3-031-19815-1_40}
}

@inproceedings{sclip,
  title        = {{SCLIP}: Rethinking Self-Attention for Dense Vision-Language
                  Inference},
  author       = {Wang, Feng and Mei, Jieru and Yuille, Alan L.},
  booktitle    = {Computer Vision -- ECCV 2024},
  series       = {Lecture Notes in Computer Science},
  volume       = {15079},
  pages        = {315--332},
  publisher    = {Springer},
  year         = {2024},
  doi          = {10.1007/978-3-031-72664-4_18}
}

@inproceedings{gem,
  title        = {Grounding Everything: Emerging Localization Properties in
                  Vision-Language Transformers},
  author       = {Bousselham, Walid and Petersen, Felix and Ferrari, Vittorio and
                  Kuehne, Hilde},
  booktitle    = {Proceedings of the IEEE/CVF Conference on Computer Vision and
                  Pattern Recognition (CVPR)},
  pages        = {3828--3837},
  month        = jun,
  year         = {2024}
}

@inproceedings{clearclip,
  title        = {{ClearCLIP}: Decomposing {CLIP} Representations for Dense
                  Vision-Language Inference},
  author       = {Lan, Mengcheng and Chen, Chaofeng and Ke, Yiping and
                  Wang, Xinjiang and Feng, Litong and Zhang, Wayne},
  booktitle    = {Computer Vision -- ECCV 2024},
  series       = {Lecture Notes in Computer Science},
  volume       = {15105},
  pages        = {143--160},
  publisher    = {Springer},
  year         = {2024},
  doi          = {10.1007/978-3-031-72970-6_9}
}

@inproceedings{proxyclip,
  title        = {{ProxyCLIP}: Proxy Attention Improves {CLIP} for
                  Open-Vocabulary Segmentation},
  author       = {Lan, Mengcheng and Chen, Chaofeng and Ke, Yiping and
                  Wang, Xinjiang and Feng, Litong and Zhang, Wayne},
  booktitle    = {Computer Vision -- ECCV 2024},
  series       = {Lecture Notes in Computer Science},
  volume       = {15126},
  pages        = {70--88},
  publisher    = {Springer},
  year         = {2024},
  doi          = {10.1007/978-3-031-73113-6_5}
}

\end{document}